\documentclass[10pt]{article}

\newif\ifanon
\anonfalse

\usepackage[preprint]{tmlr}

\usepackage[utf8]{inputenc}

\usepackage{amsmath}
\usepackage{amssymb}   
\usepackage{amsthm}
\usepackage{booktabs}  
\usepackage{graphicx}  

\usepackage{hyperref}
\usepackage{url}
\hypersetup{colorlinks=true, linkcolor=black, citecolor=black, urlcolor=blue}

\newtheorem{proposition}{Proposition}
\newtheorem{corollary}{Corollary}
\newtheorem{lemma}{Lemma}

\title{A Feasibility Condition for Confidence-Gated Tiered LLM Inference}

\author{\name Sumit P. Ahuja \email sumit@mainlabs.ai \\
        \addr Main Labs}

\def\month{MM}   
\def\year{YYYY}  
\def\openreview{\url{https://openreview.net/forum?id=XXXX}}

\begin{document}

\maketitle

\begin{abstract}
Tiered inference distributes large language model computation across a hierarchy
of devices, from on-device and edge hardware to the datacenter, and is
increasingly offered as the route to cheaper, lower-latency, and more private
serving. Most such systems orchestrate frozen model checkpoints, and most decide
what to resolve locally and what to escalate using a confidence signal produced
by a forward-deployed model. We give a closed-form condition for when no
threshold on that signal makes a tiered cascade worthwhile relative to sending
every query to the backstop. Modeling the routing
decision as a binary test and separating local from escalation cost through an
architecture cost ratio $\kappa$, we derive a feasibility condition bracketed by
a signal-free floor, below which a cascade is feasible at any signal quality, and
a perfect-signal ceiling, above which none is. Read along the $\kappa$ axis the
condition inverts: where local compute is billed, cost binds, matching a recent
empirical finding; where it is sunk, on decentralized hardware, feasibility becomes
unconditional and signal quality instead sets what the cascade achieves. We turn the condition into three design
requirements that existing trained-in primitives satisfy only in isolation, and
we propose, as future work, a co-designed architecture and a protocol to test it.
The condition and the inversion hold independently of that architecture; they are
the paper's contribution.
\end{abstract}


\section{Introduction}
\label{sec:introduction}

Large language model inference is concentrated in a small number of
capital-intensive datacenters. The arrangement is strained along several axes
at once: the marginal cost of frontier-model queries at scale, the energy and
capital intensity of the serving fleet, wide-area latency to centralized
endpoints, and data-residency rules that prohibit some traffic from leaving its
jurisdiction. A substantial body of work responds by pushing inference outward,
distributing computation across a tiered hierarchy that runs from on-device and
edge hardware, through regional points of presence, to the datacenter. From the
standpoint of distributed-systems design, the question this raises is not
whether such tiering reduces some costs (it plainly can) but whether it
relocates the constraint that actually binds, or only relocates the compute. We
argue that the answer turns on a single quantity the existing literature has
left largely implicit. We approach the problem as one of decentralized-systems
design rather than model serving\ifanon\else, a standpoint grounded in prior work
on decentralized data infrastructure \citep{AhujaZeroParty}\fi; from it, the
distinction between relocating the constraint that actually binds and relocating
the compute is the natural one to draw\ifanon.\else, and the framing that follows was arrived
at independently of the concurrent results we engage in
Section~\ref{sec:gap}.\fi

Four otherwise-separate families of work make up the current toolkit for tiered
inference. Pipeline-distributed serving splits a single model's layers across
geographically separated nodes \citep{Borzunov2023Petals}. Prefill/decode
disaggregation separates the compute-bound prefill phase from the
memory-bandwidth-bound decode phase onto distinct hardware, and is now standard
in production serving systems \citep{Zhong2024DistServe, Patel2024Splitwise,
Qin2024Mooncake}. Model routing and cascades resolve easy queries with a small
model and escalate hard ones to a larger model \citep{Chen2023FrugalGPT,
Ding2024HybridLLM, Ong2024RouteLLM}. Cross-tier speculative decoding drafts
tokens on a forward tier and verifies them upstream in batch. All four
orchestrate frozen model checkpoints rather than re-engineering the model. Three
of them, routing and cascades, cross-tier speculative decoding, and
confidence-gated pipeline escalation, decide what to escalate using a
confidence signal produced by a forward-deployed model. Phase disaggregation is
the exception: it splits by computational phase, not by query difficulty, and it
is the one form of tiering that has reached production at scale. The contrast is
instructive. The tiering that sidesteps the confidence signal has consolidated
into a production standard; the tiering that depends on it, though widely
deployed, persists as a tuning problem solved afresh in each deployment.

That signal has been studied extensively, and a recurring finding is that it is
poorly behaved: token-level confidence is miscalibrated and sensitive to prompt
wording, verbalized confidence is overconfident, and the one robust alternative,
sampling-based self-consistency, is expensive \citep{Wen2025KnowYourLimits,
Kotte2026UCCI}. The usual response is to treat the deferral threshold as a
hyperparameter tuned per deployment \citep{Chen2023FrugalGPT}, or to patch the
signal with post-hoc calibration or a learned router
\citep{Zellinger2025EarlyAbstention}. Each patch concedes the premise that the
raw signal is inadequate without asking the prior question: under what
conditions can any threshold on such a signal make a cascade worthwhile at all?
That question has a formal answer, which to our knowledge has not been stated.

We give it. Modeling the routing decision as a binary test characterized by its
operating point on a receiver-operating-characteristic curve, and denominating
local and escalation cost separately through an architecture cost ratio
$\kappa$ (the marginal cost of a local forward pass divided by the cost of an
escalation), we derive a closed-form condition under which the feasible set of
routing thresholds is empty: no threshold yields a cascade that beats sending
every query to the datacenter. The condition couples two quantities the signal
literature treats separately, the router's discrimination and the router's
cost, and shows they move in opposite directions, so that improving one can
vacate the feasible region through the other. Concurrent work reaches a closely
related conclusion empirically. \citet{Bouchard2026Escalation} characterizes
the cascade cost--quality frontier and finds that cascade performance is limited
primarily by structural cost rather than by routing-signal strength. We build on
that finding directly and locate its scope.
The structural-cost conclusion holds where local compute is a billed cost, the
cloud-API regime, in which $\kappa$ is appreciable. On sunk-cost decentralized
hardware, where the marginal cost of a local forward pass approaches zero and
$\kappa \to 0$, feasibility becomes unconditional (Section~\ref{sec:inequality})
and signal quality instead determines the operating point the cascade achieves.
On this account, decentralization alters which lever moves the outcome, rather
than only reducing cost.

The condition is diagnostic but not only diagnostic. It implies necessary
requirements for any architecture meant to cross the ceiling rather than tune
beneath it: a calibrated escalation signal available at near-zero marginal cost,
a cross-tier payload compressible below the available bandwidth, and placement
that respects tier locality. Each requirement is already met in isolation by a
trained-in primitive, calibrated early exit \citep{CECoLLM2024,
Schuster2022CALM}, learned activation bottlenecks \citep{Matsubara2022BottleFit},
and locality-aware routing objectives \citep{Li2024LocMoEPlus}, yet no current
model meets all three jointly. We frame the unified, tier-native co-design that
would as a research program and outline a protocol to test it. We do not
implement or evaluate that architecture here.\ifanon\else{} This is a position
paper, and the architecture is its forward-looking conclusion rather than a
validated result.\fi

\paragraph{Contributions.}
\begin{itemize}
  \item \textbf{A unifying critique.} We show that four families of tiered LLM
  inference share a single property, deployment-layer orchestration of frozen
  checkpoints, and isolate the confidence-signal dependence common to the
  routing, cascading, and cross-tier speculative-decoding forms, and tellingly
  absent from the one form, phase disaggregation, that has reached production
  (Section~\ref{sec:mechanisms}).
  \item \textbf{A feasibility condition for tiered cascades.} We introduce the
  architecture cost ratio $\kappa$ and derive a closed-form condition under which
  the feasible set of routing thresholds is empty, a conditional-impossibility
  result, together with a regime-scaled budget $N^\star$ bounding the
  self-consistency samples a feasible cascade can afford
  (Section~\ref{sec:inequality}).
  \item \textbf{Design requirements and a primitive gap.} We show the condition
  implies three necessary requirements for an architecture that escapes the
  ceiling, and we map existing trained-in primitives onto them, establishing that
  each is satisfied in isolation and none jointly
  (Section~\ref{sec:requirements}).
  \item \textbf{A regime inversion and a research program.} We bound the
  structural-cost conclusion of \citet{Bouchard2026Escalation} to the
  paid-compute regime, identify the sunk-cost decentralized regime in which
  signal quality alone determines the achievable operating point, and specify a
  candidate co-designed architecture with
  a validation protocol as future work
  (Sections~\ref{sec:requirements}--\ref{sec:architecture}).
\end{itemize}

\noindent
Section~\ref{sec:background} fixes terminology and the tier model.
Section~\ref{sec:mechanisms} develops the unifying critique.
Sections~\ref{sec:gap}--\ref{sec:propositions} state the gap and our claims as
falsifiable propositions. Section~\ref{sec:inequality} derives the feasibility
condition and the regime inversion. Section~\ref{sec:requirements} turns the
condition into design requirements, and Section~\ref{sec:architecture} states the
candidate architecture and its validation protocol. The remaining sections cover
discussion, limitations, ethics, and future work.


\section{Background and definitions}
\label{sec:background}

This section fixes the objects the rest of the paper reasons about: the tier
model and the boundary we call the seam, the two phases of inference, the
selective-prediction machinery that supplies the routing decision, and the cost
ratio that distinguishes deployment regimes. Notation is collected in
Table~\ref{tab:notation}.

\subsection{The tier model and the seam}
\label{sec:background:tiers}

We model tiered inference as a chain of compute tiers ordered by capability and
distance from the user: a \emph{local} tier (the user's own device), one or more
\emph{edge} and \emph{regional} tiers, and a \emph{datacenter} (DC) tier hosting
the most capable model. We refer to any tier below the datacenter as a
\emph{forward tier} and to the datacenter model as the \emph{backstop}. A query
is resolved at the lowest tier whose output is accepted; otherwise it is
\emph{escalated} upstream. The \emph{seam} is the boundary between a forward tier
and the next tier up: the point at which a routing decision is taken and across
which an escalated query, or an intermediate representation of it, must travel.

Tiers differ not only in capability but in \emph{who owns the compute}, and this
distinction is central to our analysis. On rented or metered infrastructure, a
cloud API, a per-token endpoint, a forward-tier forward pass carries a billed
marginal cost. On hardware the operator already owns, an edge node, an
on-premises accelerator, a user device, the capital is sunk and the marginal
cost of an additional forward pass approaches zero. We treat this as a property
of the deployment, not of the model, and make it explicit through the cost ratio
defined in Section~\ref{sec:background:cost}.

A forward-tier model is typically a \emph{small language model} (SLM). We adopt
the working convention of \citet{Belcak2025SLMAgentic}, who, as of 2025, regard
models below roughly ten billion parameters as small; nothing in our argument
depends on a sharp threshold.

\subsection{Inference phases}
\label{sec:background:phases}

Autoregressive inference has two phases with distinct resource profiles. The
\emph{prefill} phase ingests the prompt and is compute-bound; the \emph{decode}
phase generates tokens one at a time and is bound by memory bandwidth, since each
step rereads the model weights and the growing key--value cache
\citep{Zhong2024DistServe, Patel2024Splitwise}. Two consequences matter here.
First, phase disaggregation, assigning prefill and decode to separate hardware, is a form of tiering that splits by phase rather than by query difficulty, and
so does not consult a confidence signal. Second, any routing signal that requires
generating one or more full responses (Section~\ref{sec:background:selective})
incurs decode cost proportional to the number of responses generated, which is
the mechanism by which a more reliable signal can become an expensive one.

\subsection{Selective prediction and the routing decision}
\label{sec:background:selective}

The decision to resolve a query locally or escalate it is an instance of
\emph{selective prediction}: a model abstains, or defers, rather than commit to a
prediction it is unlikely to get right. The framework originates in selective
classification with a reject option \citep{Geifman2017Selective,
Geifman2019SelectiveNet} and in learning to defer to an expert
\citep{Mozannar2020Defer}; tiered inference instantiates it with the backstop in
the role of the expert.

A routing rule thresholds a \emph{confidence signal}. We group signals by the
decode cost they incur. \emph{Cheap} signals are read from a single forward pass:
the sequence log-probability, token entropy, or the margin between the top two
token probabilities. \emph{Verbalized} signals prompt the model to state a
confidence and cost roughly one additional pass. The \emph{robust} alternative,
\emph{self-consistency}, samples $N$ responses and measures their agreement, at a
cost that scales with $N$. A recurring empirical finding, which we take as given
rather than re-establish, is that the cheap and verbalized signals are
miscalibrated and brittle, while the signal that discriminates reliably is the
costly one \citep{Wen2025KnowYourLimits, Kotte2026UCCI}.

For the analysis in Section~\ref{sec:inequality} we abstract the routing rule to a
binary test of whether the forward model has \emph{erred} on a query. Let $q$ be
the forward model's accuracy on the workload and $u$ the backstop's error rate.
The test is characterized by two rates: its \emph{sensitivity}
$s = \Pr(\text{escalate}\mid\text{forward model wrong})$, the rate at which it
catches forward-tier errors, and its \emph{false-escalation rate}
$f = \Pr(\text{escalate}\mid\text{forward model correct})$. The pair $(f,s)$ is an
operating point on a receiver-operating-characteristic (ROC) curve; a perfect
router sits at $(0,1)$ and a signal carrying no information sits on the diagonal
$s=f$. The induced \emph{escalation rate} is
\begin{equation}
  e \;=\; s\,(1-q) \;+\; f\,q .
  \label{eq:escalation-rate}
\end{equation}
The same trade-off is reported as a \emph{risk--coverage} curve, where coverage is
the fraction of queries answered locally ($1-e$) and risk is the error rate among
them; its area summarizes a signal's selective-prediction quality. We write
$\delta$ for the excess error over the backstop baseline the operator is willing
to tolerate in exchange for answering locally.

\subsection{Cost and the architecture ratio}
\label{sec:background:cost}

Let $c_L$ be the marginal cost of resolving a query on the forward tier and $c_D$
the marginal cost of an escalation to the backstop, both in the operator's
binding currency (money, energy, or throughput). We define the \emph{architecture
cost ratio}
\begin{equation}
  \kappa \;=\; \frac{c_L}{c_D},
  \label{eq:kappa}
\end{equation}
the cost of a local forward pass relative to an escalation. As argued in
Section~\ref{sec:background:tiers}, $\kappa$ is set by the deployment: appreciable
when forward-tier compute is billed (cloud-API cascades), and approaching zero
when it is sunk (owned edge hardware). A self-consistency signal of $N$ samples
multiplies the local cost, $c_L \mapsto N\,c_L$, and therefore scales $\kappa$ by
$N$. Section~\ref{sec:inequality} shows that whether a cascade is feasible at all,
and which of signal quality or signal cost limits what it achieves, is governed
by $\kappa$ together with the router's operating point.

\begin{table}[t]
  \centering
  \begin{tabular}{ll}
    \toprule
    Symbol & Meaning \\
    \midrule
    $q$ & accuracy of the forward (small) model on the workload \\
    $u$ & error rate of the datacenter backstop model \\
    $\delta$ & tolerated excess error over the backstop baseline \\
    $s$ & router sensitivity, $\Pr(\text{escalate}\mid\text{wrong})$ \\
    $f$ & router false-escalation rate, $\Pr(\text{escalate}\mid\text{correct})$ \\
    $e$ & escalation rate, $s(1-q)+fq$ \\
    $c_L,\,c_D$ & marginal cost of a local pass; of an escalation \\
    $\kappa$ & architecture cost ratio, $c_L/c_D$ \\
    $N$ & number of self-consistency samples \\
    \bottomrule
  \end{tabular}
  \caption{Notation used throughout. Section~\ref{sec:inequality} derives the
  feasibility condition in these terms.}
  \label{tab:notation}
\end{table}

\begin{figure}[t]
  \centering
  \includegraphics[width=\linewidth]{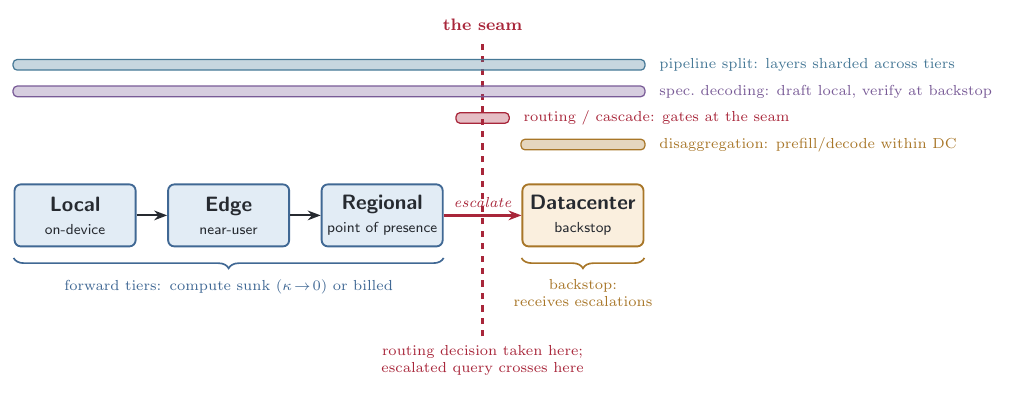}
  \caption{The tier model and the seam. Forward tiers (local, edge, regional)
  resolve queries that a routing decision at the seam accepts; the datacenter
  backstop receives escalations. The architecture cost ratio $\kappa$
  (Section~\ref{sec:background:cost}) is small where forward-tier compute is
  sunk and appreciable where it is billed.}
  \label{fig:tier-model}
\end{figure}


\section{Four mechanisms, one fault line}
\label{sec:mechanisms}

The current toolkit for tiered inference comprises four families of work. They
are usually presented separately, as systems contributions with different
benchmarks and different vocabularies. We organize them instead by a single
question (does the mechanism consult a confidence signal to decide what crosses the seam?) because the answer, and not the systems detail, predicts which
mechanisms have consolidated into production standards and which remain a tuning
problem. The four are summarized in Table~\ref{tab:mechanisms}.

\paragraph{Pipeline-distributed serving.}
A single model's layers are partitioned across nodes, so that a forward pass
traverses the network: tokens enter at the first shard and activations are
relayed shard to shard. Petals \citep{Borzunov2023Petals} runs this over
volunteer hardware across the wide-area network; subsequent work optimizes block
placement and request routing for geographically distributed deployments
\citep{GeoDistInference2025}. Every query runs every layer; there is no
difficulty assessment and no escalation decision. What crosses the seam is the
full activation tensor at each partition boundary, on every token. The mechanism
operates on a frozen checkpoint and consults no confidence signal. Its limiting
cost is communication, and it has remained a niche deployment rather than a
production default.

\paragraph{Phase disaggregation.}
Prefill and decode are assigned to separate hardware pools, eliminating the
interference that arises when the compute-bound and bandwidth-bound phases share a
device \citep{Zhong2024DistServe, Patel2024Splitwise}. What crosses the seam is
the key--value cache, transferred once per request from the prefill pool to the
decode pool, a cost driven below significance by intra-datacenter interconnect
\citep{Qin2024Mooncake}. The split is by computational phase, not by query
difficulty; no confidence signal is read. Disaggregation operates on a frozen
checkpoint and has been adopted as a default across production serving frameworks.

\paragraph{Routing and cascades.}
A small forward model attempts each query; a routing rule decides whether to
accept its answer or escalate to a larger model. Cascades escalate after
generating a candidate and judging it inadequate \citep{Chen2023FrugalGPT,
Aggarwal2023AutoMix}; predictive routers decide before generation from features of
the query \citep{Ding2024HybridLLM, Ong2024RouteLLM}. What crosses the seam is the
query itself, and only for the fraction escalated. The decision is a threshold on
a confidence signal, and the central engineering problem is choosing that
threshold: it is tuned per deployment, and supplemented with post-hoc calibration
or a learned router when the raw signal proves inadequate \citep{Kotte2026UCCI,
Zellinger2025EarlyAbstention}. The mechanism operates on frozen checkpoints and is
wholly dependent on the confidence signal.

\paragraph{Cross-tier speculative decoding.}
The forward tier drafts tokens and the backstop verifies them, exploiting the
gap between the cost of generating a token and the cost of checking one. In its
basic form every drafted token is verified, gated by the backstop's distribution
rather than by a forward-tier signal. Its collaborative edge--cloud variants are
selective: the forward model resolves high-confidence tokens locally and escalates
only the uncertain ones, reintroducing a confidence threshold and, with it, the
round-trip latency of stop-and-wait coordination \citep{DistSpecDecode2025,
EdgeCloudSpecDecode2026}. What crosses the seam is a stream of draft tokens and
verifications. The mechanism operates on frozen checkpoints; its selective
variants depend on the confidence signal, its basic form does not.

All four families share a forward-tier model that is small by the working
convention of \citet{Belcak2025SLMAgentic}, and all four orchestrate frozen
checkpoints: none re-engineers the model so that its parameters are aware of the
tier boundary they will be split across. That shared property is the first half of
the critique. The second half is the division visible in the last column of
Table~\ref{tab:mechanisms}. Two mechanisms, pipeline-distributed serving and
phase disaggregation, split the computation \emph{structurally}, by layer or by
phase, and consult no confidence signal. Two, routing/cascades and the selective
form of cross-tier speculative decoding, split it by \emph{difficulty}, and
stake their correctness on a forward model's estimate of its own reliability.

\begin{table}[t]
  \centering
  \small
  \begin{tabular}{@{}lllcc@{}}
    \toprule
    Mechanism & Tier scope & Crosses the seam & Frozen & Signal-gated \\
    \midrule
    Pipeline-distributed     & spans tiers          & layer activations, every token & yes & no \\
    Phase disaggregation     & within datacenter    & KV cache, once per request     & yes & no \\
    Routing / cascades       & forward $\to$ backstop & the query, when escalated    & yes & yes \\
    Cross-tier spec.\ decoding & forward $\to$ backstop & draft tokens + verification & yes & partial \\
    \bottomrule
  \end{tabular}
  \caption{Four families of tiered inference, organized by whether a confidence
  signal gates what crosses the seam. All operate on frozen checkpoints; the
  signal dependency, not the systems detail, separates the structural splits
  (top two) from the difficulty-routed forms (bottom two).}
  \label{tab:mechanisms}
\end{table}

This division tracks a difference in outcome. The structural splits have
consolidated: disaggregation is a production default, and the layer-split form is
limited only by communication, a quantity that does not depend on the model's
self-knowledge. The difficulty-routed forms have not consolidated in the same
sense. They are deployed, but each deployment re-solves the same problem, selecting and maintaining a threshold on a signal that drifts with the workload, and the literature that supports them is, to a first approximation, a catalogue of
methods for propping up that signal. The mechanism that avoids the confidence
signal scales with bandwidth; the mechanism that depends on it scales with the
quality of an estimate that the forward model is, by construction, not reliably
able to make. The confidence signal is the fault line.

We are not the first to enumerate these methods. A recent survey catalogues
routing and cascading techniques and their reported trade-offs in detail
\citep{MoslemKelleher2026RoutingSurvey}. Our purpose is different: not to catalogue
the methods but to locate the dependency they share, and to ask what that
dependency costs in the limit. The next section states the gap that question
exposes, and Section~\ref{sec:inequality} answers it with a feasibility condition
that holds regardless of how the threshold is chosen.


\section{The gap: a shared dependency, uncharacterized}
\label{sec:gap}

The confidence signal has been studied from two directions, and both take the
dependency on it as given. From below, the selective-prediction and abstention
literature measures how well the signal separates correct outputs from incorrect
ones, and documents that it does so poorly: confidence is miscalibrated, brittle
to prompting, and reliable only when made expensive \citep{Geifman2017Selective,
Guo2017Calibration, Wen2025KnowYourLimits}. From above, the systems literature
treats the deferral threshold as the object to optimize, tuning it per
deployment \citep{Chen2023FrugalGPT}, modeling the cascade probabilistically to
set it \citep{Zellinger2025RationalTuning}, or replacing the raw signal with a
calibrated or learned one \citep{Kotte2026UCCI, Zellinger2025EarlyAbstention}.
Both directions improve the operating point. Neither asks whether an operating
point that clears the bar exists at all.

The nearest step toward that question is recent. \citet{Bouchard2026Escalation}
characterizes the cascade cost--quality frontier and finds, across the settings
studied, that cascade performance is limited primarily by structural cost rather
than by the strength of the routing signal. This is a characterization rather than
another method, and we build on it directly. But it is a characterization at a
fixed location in cost space: the analysis reads off the binding constraint in the
regime where local computation is a billed cost. Whether that constraint is
intrinsic to tiered cascades or an artifact of the cost regime, and what becomes
of it when the marginal cost of a local forward pass goes to zero, is left open.

The gap is therefore narrow and specific. What is missing, to our knowledge, is a
closed-form condition, expressed in the quantities the selective-prediction
literature already measures: the router's sensitivity and false-escalation rate,
the two tiers' error rates, and the tolerated excess error, stating when no
threshold makes the cascade worthwhile, together with the dependence of that
condition on the architecture cost ratio $\kappa$. Such a condition would do two
things no existing result does. It would let an operator decide, \emph{before}
tuning a threshold, whether any threshold can succeed on a given workload and cost
structure. And by treating $\kappa$ as an explicit axis rather than a fixed
backdrop, it would reveal whether signal quality or signal cost is the limiting
quantity, and so whether decentralization, which drives $\kappa$ toward zero,
relaxes the ceiling or merely slides along it. Section~\ref{sec:propositions}
states the argument this exposes as falsifiable propositions, and
Section~\ref{sec:inequality} derives the condition.


\section{Claims}
\label{sec:propositions}

We state our claims as four propositions, ordered by what this paper does with
each. The first we argue from the structure of existing mechanisms; the next two
we derive; the last we hypothesize and specify a test for. Each is stated so that
a concrete result would falsify it.

\paragraph{P1 (Dependency).}
The quality--cost trade-off achievable by difficulty-routed tiering is governed by
the forward model's confidence signal, concretely, by the router's operating
point $(f,s)$ on its ROC curve. System-level engineering that does not improve
that operating point cannot move the frontier beyond what the signal admits.
\emph{Support:} Section~\ref{sec:mechanisms}, where every difficulty-routed
mechanism is shown to reduce to a threshold on this signal. \emph{A falsifier:} a
difficulty-routed system whose quality--cost frontier is insensitive to controlled
degradation of the confidence signal.

\paragraph{P2 (Ceiling).}
There exist workloads and cost structures, combinations of forward-model accuracy
$q$, backstop error $u$, tolerated excess error $\delta$, and architecture cost
ratio $\kappa$, for which the feasible set of routing thresholds is empty: no
threshold yields a cascade that beats sending every query to the backstop. The
ceiling is conditional, not universal; it binds in an identifiable region of the
parameter space. \emph{Support:} Section~\ref{sec:inequality}, where the region is
derived in closed form. \emph{A falsifier:} a threshold achieving net benefit on a
workload and cost structure the condition places in the empty region.

\paragraph{P3 (Inversion).}
Which quantity governs the outcome is regime-dependent through $\kappa$. Where
$\kappa$ is appreciable, local computation billed, as in cloud-API cascades, the
cost constraint shrinks the feasible set and above $\kappa^\star_{\mathrm{ceil}}$
empties it, so signal improvements have bounded value and, past the ceiling, none;
this is the regime in which \citet{Bouchard2026Escalation} finds structural cost
dominant. As $\kappa \to 0$, local computation sunk, as on owned edge hardware,
feasibility becomes unconditional (Corollary~\ref{cor:floor}) and signal quality
instead determines the achievable operating point: the locally resolved fraction
$1 - e_q$. Marginal analysis inverts accordingly: at large $\kappa$ only
cost-structure changes move the outcome; at small $\kappa$ only signal improvements
do. Decentralization therefore changes which lever moves the outcome, not merely the
magnitude of cost. \emph{Support:} Section~\ref{sec:inequality}.
\emph{A falsifier:} a demonstration that signal improvements fail to move the
achieved operating point on genuinely sunk-cost hardware, or that cost-structure
changes continue to dominate the outcome as $\kappa \to 0$.

\paragraph{P4 (Co-design escape).}
The ceiling is a property of frozen-checkpoint orchestration, not of tiering as
such. A model co-designed so that its parameters are aware of the tier boundary
they are split across, one that jointly meets the three requirements of
Section~\ref{sec:requirements}, can reach operating points unattainable by
thresholding a frozen model's signal. This proposition is a hypothesis, not a
result: we do not build or evaluate such a model here. \emph{Test:} the protocol
specified in Section~\ref{sec:architecture}. \emph{A falsifier:} a demonstration
that the three requirements cannot be met jointly, or that a model meeting them
fails to improve the quality--cost frontier over a tuned frozen-checkpoint cascade.

\medskip
\noindent
P2 and P3 are the formal core, and the remainder of the paper turns on them.
Section~\ref{sec:inequality} derives both from the definitions of
Section~\ref{sec:background}.


\section{The feasibility condition}
\label{sec:inequality}

We now make the dependency of Section~\ref{sec:mechanisms} precise. Modeling the
router as the binary test of Section~\ref{sec:background:selective}, we write the
cascade's cost and error in the notation of Table~\ref{tab:notation} and ask when
some operating point $(f,s)$ satisfies both a cost and a quality requirement. The
answer is a closed-form condition on the architecture cost ratio $\kappa$ and the
router's discrimination. It is empty in an identifiable region, no threshold
makes the cascade worthwhile (P2), and the region's location moves with $\kappa$
in a way that inverts which lever moves the outcome (P3).

\subsection{Two constraints}
\label{sec:inequality:constraints}

We analyze the generate-then-judge cascade: every query is run on the forward
model $L$ at cost $c_L$, a routing test reads a confidence signal from that run,
and the escalated fraction is additionally run on the backstop $D$ at cost $c_D$.
The escalation rate is $e = s(1-q) + fq$ (Eq.~\ref{eq:escalation-rate}). We compare
against the baseline of sending every query to $D$, at cost $c_D$ per query.

\paragraph{Cost.}
The cascade's per-query cost is $c_L + e\,c_D$. It beats the baseline when
\begin{equation}
  c_L + e\,c_D < c_D
  \;\;\Longleftrightarrow\;\;
  e < 1 - \kappa,
  \label{eq:cost-constraint}
\end{equation}
dividing by $c_D$ and using $\kappa = c_L/c_D$ (Eq.~\ref{eq:kappa}).

\paragraph{Quality.}
Each query falls in one of four cells according to whether $L$ is correct and
whether the router escalates. A kept query takes $L$'s answer; an escalated query
takes $D$'s. Summing the error contributions,
\begin{align}
  E_{\text{cascade}}
  &= \underbrace{q(1-f)\cdot 0}_{\text{$L$ right, kept}}
   + \underbrace{qf\cdot u}_{\text{$L$ right, escalated}}
   + \underbrace{(1-q)(1-s)\cdot 1}_{\text{$L$ wrong, kept}}
   + \underbrace{(1-q)s\cdot u}_{\text{$L$ wrong, escalated}} \notag\\
  &= (1-q)(1-s) + u\,\big[\,qf + (1-q)s\,\big]
   = (1-q)(1-s) + u\,e .
  \label{eq:cascade-error}
\end{align}
Tolerating excess error $\delta$ over the backstop baseline $u$ gives the quality
requirement
\begin{equation}
  (1-q)(1-s) + u\,e \;\le\; u + \delta .
  \label{eq:quality-constraint}
\end{equation}
We assume throughout the non-degenerate regime $0 < \delta < (1-q) - u$, i.e.\ the
forward model's error rate exceeds the backstop's ($\,(1-q) > u$, so escalation can
reduce error) and the tolerance is smaller than the gap to close (so the cascade is
not trivially satisfied by $L$ alone). We also take $u$ independent of which queries
escalate; Section~\ref{sec:limitations} discusses relaxing this, which strengthens
the result.

\subsection{When no threshold suffices}
\label{sec:inequality:ceiling}

Let $e_q$ denote the smallest escalation rate at which the router, operating at its
best point for that rate, satisfies the quality requirement
\eqref{eq:quality-constraint}. A better-discriminating router catches $L$'s errors
with fewer escalations and so has a smaller $e_q$.

\begin{proposition}[Feasibility]
\label{prop:feasibility}
A cascade admits an operating point satisfying both \eqref{eq:cost-constraint} and
\eqref{eq:quality-constraint} if and only if
\begin{equation}
  \kappa \;<\; 1 - e_q .
  \label{eq:feasibility}
\end{equation}
\end{proposition}

\noindent\emph{Proof.}
Quality requires $e \ge e_q$ by definition of $e_q$; cost requires $e < 1-\kappa$ by
\eqref{eq:cost-constraint}. A common $e$ exists iff $e_q < 1-\kappa$, i.e.\ iff
$\kappa < 1 - e_q$. \hfill$\square$

The content of Proposition~\ref{prop:feasibility} is that $e_q$ is set by signal
quality while $1-\kappa$ is set by cost structure, and feasibility requires the
first to fit under the second. The two bracket a usable range of $\kappa$, obtained
by evaluating $e_q$ at the extremes of router quality.

\begin{corollary}[Signal-free floor; conditional impossibility]
\label{cor:floor}
For an uninformative signal (the ROC diagonal $s=f$, so $e=s$), quality holds iff
$s \ge 1 - \delta/[(1-q)-u]$, giving $e_q = 1 - \delta/[(1-q)-u]$ and
\begin{equation}
  \kappa \;<\; \kappa^\star_{\mathrm{floor}} \;=\; \frac{\delta}{(1-q)-u}.
  \label{eq:floor}
\end{equation}
Equivalently, when $\kappa\,[(1-q)-u] \ge \delta$, no cascade driven by an
uninformative signal is feasible: feasibility then \emph{requires} a strictly
informative signal.
\end{corollary}

\begin{corollary}[Perfect-signal ceiling; absolute impossibility]
\label{cor:ceiling}
For a perfectly discriminating signal ($f=0$, escalating only $L$'s errors), quality
holds at $e_q = [(1-q)-u-\delta]/(1-u)$, giving
\begin{equation}
  \kappa \;<\; \kappa^\star_{\mathrm{ceil}} \;=\; 1 - \frac{(1-q)-u-\delta}{1-u}.
  \label{eq:ceiling}
\end{equation}
For $\kappa \ge \kappa^\star_{\mathrm{ceil}}$ no router, however good, makes the
cascade worthwhile. This is the conditional impossibility of P2.
\end{corollary}

\noindent
Between the two thresholds, $\kappa^\star_{\mathrm{floor}} \le \kappa <
\kappa^\star_{\mathrm{ceil}}$, feasibility is possible but contingent on signal
quality: it holds only if the available signal is informative enough to push $e_q$
below $1-\kappa$. This middle band is where deployed cascades operate, and it is
the band the rest of the paper is about.

\subsection{The inversion}
\label{sec:inequality:inversion}

Read Proposition~\ref{prop:feasibility} along the $\kappa$ axis
(Figure~\ref{fig:feasibility}). As $\kappa \to 0$, local computation sunk, the
decentralized-edge regime, the cost bound $1-\kappa \to 1$; and once
$\kappa < \kappa^\star_{\mathrm{floor}}$, \eqref{eq:feasibility} holds for every
signal (Corollary~\ref{cor:floor}): feasibility becomes unconditional. What remains
signal-limited is the operating point. Cost-minimizing operation escalates at the
quality-minimal rate $e = e_q$, so the fraction resolved locally, $1 - e_q$, is set
by discrimination alone: a chance signal is feasible below the floor but must
escalate nearly everything ($e_q \approx 0.94$ at the worked point of
Section~\ref{sec:inequality:worked}), while a discriminating signal keeps the
workload local. Cost stops excluding operating points; signal quality prices them.
As $\kappa$ grows, local computation billed, the cloud-API regime, the cost bound
tightens until, past $\kappa^\star_{\mathrm{ceil}}$, no signal can satisfy both
constraints: cost binds absolutely. What governs the outcome thus inverts across the $\kappa$ axis. The
regime in which \citet{Bouchard2026Escalation} finds structural cost dominant is the
upper part of this axis; the decentralized regime sits at the lower end, where the
same analysis makes signal quality the quantity that sets the outcome.
Decentralization does not loosen a fixed ceiling: it moves the system to the part of
the axis where feasibility is free and value is signal-priced (P3).

\subsection{The price of a robust signal}
\label{sec:inequality:selfconsistency}

The one signal that reliably beats the diagonal is self-consistency, which samples
$N$ responses and scores their agreement (Section~\ref{sec:background:selective}).
It multiplies the local cost, $c_L \mapsto N c_L$, hence $\kappa \mapsto N\kappa$.
Substituting into \eqref{eq:feasibility}, the affordable sample budget is
$N^\star = (1-e_q)/\kappa$. Taking the signal-free floor as the reference point
(the conservative case, where the samples are needed merely to reach usable
discrimination),
\begin{equation}
  N^\star \;=\; \frac{\delta}{\kappa\,[(1-q)-u]} \;=\; \frac{\delta\,\rho}{(1-q)-u},
  \qquad \rho \equiv \frac{1}{\kappa} = \frac{c_D}{c_L}.
  \label{eq:nstar}
\end{equation}
For $q=0.6$, $u=0.05$, $\delta=0.02$, and an escalation costing $\rho=30$ times a
single local pass, $N^\star = (0.02)(30)/0.35 \approx 1.7$. Fewer than two samples
exhaust that bound. Because $e_q$ itself depends on the signal the $N$ samples
purchase, the binding budget solves the fixed-point condition
$N\kappa < 1 - e_q(N)$ with $e_q(N)$ decreasing in $N$, which has no closed form;
Eq.~\eqref{eq:nstar} evaluates it at the signal-free floor and is therefore a lower
bound on the affordable budget, under the pessimistic assumption that the samples do
not improve discrimination. Evaluating instead at the perfect-signal point of
Corollary~\ref{cor:ceiling} gives the upper extreme,
$N < (1-e^{\mathrm{pf}}_q)/\kappa \approx 19.6$ for the same parameters, so at
$\rho=30$ the affordable budget lies between roughly 2 and 20 samples wherever the
cascade is feasible at all. The bracket scales linearly in $\rho = 1/\kappa$: at
$\rho=1000$, on owned hardware, the same bounds run from roughly 57 to 653, and the
budget is no longer the obstacle. The sample budget is thus a cloud-regime
constraint that dissolves as $\kappa \to 0$, the inversion of
Section~\ref{sec:inequality:inversion} arriving through a second door: where local
compute is billed, even the robust signal must fit a budget of tens of samples;
where it is sunk, signal quality alone decides.

\subsection{A worked point}
\label{sec:inequality:worked}

\begin{figure}[t]
  \centering
  \includegraphics[width=0.72\linewidth]{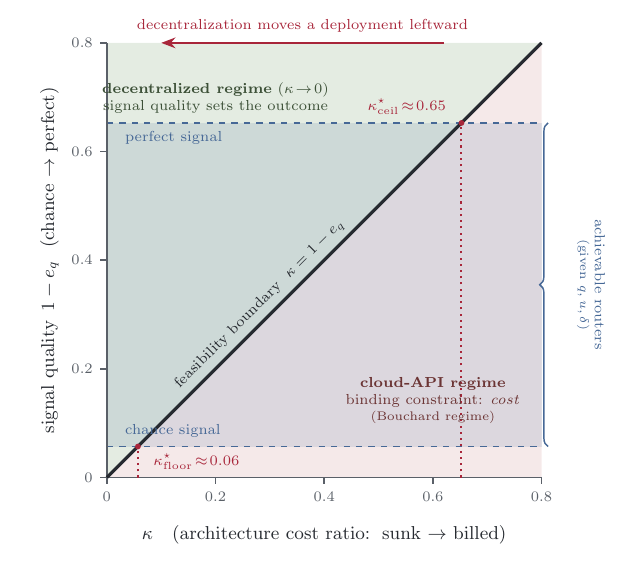}
  \caption{The feasibility region of Proposition~\ref{prop:feasibility} in the
  cost-ratio / signal-quality plane. Below $\kappa^\star_{\mathrm{floor}}$ a
  cascade is feasible at any signal quality; above $\kappa^\star_{\mathrm{ceil}}$
  it is infeasible at every signal quality; between them feasibility is contingent
  on the signal. Decentralization moves a deployment leftward, where feasibility is
  unconditional below $\kappa^\star_{\mathrm{floor}}$ and signal quality sets the
  operating point a feasible cascade achieves. Illustrative parameters $q=0.6$, $u=0.05$,
  $\delta=0.02$.}
  \label{fig:feasibility}
\end{figure}

The same parameters fix the two thresholds. With $(1-q)-u = 0.35$,
\eqref{eq:floor} gives $\kappa^\star_{\mathrm{floor}} = 0.02/0.35 \approx 0.057$ and
\eqref{eq:ceiling} gives $\kappa^\star_{\mathrm{ceil}} = 1 - 0.33/0.95 \approx 0.65$.
A cloud cascade with $\kappa \approx 0.057$--$0.65$ is in the contingent band and
lives or dies by its signal; an owned-edge deployment with $\kappa \to 0$ clears the
cost bound outright and is limited only by signal quality. Figure~\ref{fig:feasibility}
plots the feasible region in the $(\kappa, \text{signal-quality})$ plane and marks
both regimes.

The condition rests on assumptions stated where they were made: the
generate-then-judge cost model, two tiers, $u$ independent of the escalated set, and
the chance/perfect bracketing of $e_q$. Section~\ref{sec:limitations} examines each
and argues that the realistic relaxations tighten rather than loosen the ceiling.
Section~\ref{sec:requirements} turns \eqref{eq:feasibility} from a diagnostic into
design requirements.


\section{From condition to requirements}
\label{sec:requirements}

Proposition~\ref{prop:feasibility} is diagnostic: given a workload and a
deployment, it says whether any threshold can work. Read prescriptively, it also
says what an architecture must provide to enlarge the feasible region rather than
search within it. The feasibility margin $1-\kappa-e_q$ has three inputs an operator
cannot improve by tuning a threshold: the calibration of the signal (which sets
$e_q$), the marginal cost of obtaining that signal (which scales $\kappa$ under
self-consistency), and the physical cost of the escalation and the local pass
(which set $\kappa$ itself in a tiered deployment). Each yields a requirement.

\subsection{Three requirements}
\label{sec:requirements:three}

\paragraph{R1: a calibrated escalation signal at near-zero marginal cost.}
This requirement comes directly from Section~\ref{sec:inequality}. The feasibility
margin is destroyed in two ways: by a poorly calibrated signal, which raises $e_q$
toward the floor of Corollary~\ref{cor:floor}; and by buying calibration through
self-consistency, which multiplies $\kappa$ and, where $\kappa$ is appreciable,
exhausts the sample budget bounded in Eq.~\eqref{eq:nstar}. A frozen checkpoint offers one or the other: its
cheap single-pass signals are miscalibrated, and its one reliable signal is
expensive. R1 requires both at once, a signal that is calibrated \emph{and} read
at the cost of the forward pass already being computed, not a multiple of it.

\paragraph{R2: a cross-tier payload compressible below the link.}
The inequality treats the escalation cost $c_D$ as given, but in a tiered deployment
$c_D$ includes moving data across the seam. An escalation that re-runs the query
from scratch on the backstop discards the forward-tier computation already spent; an
escalation that hands off the forward tier's intermediate representation avoids that
waste only if the representation fits the cross-tier link. Pipeline-distributed
serving sits at the bad extreme of this trade-off, full activations relayed every
token (Table~\ref{tab:mechanisms}), and is bandwidth-bound as a result. R2
requires that whatever crosses the seam on escalation be compressible below the
available bandwidth, so that the handoff is cheap rather than a second bottleneck.

\paragraph{R3: placement that respects tier locality.}
The decentralized regime that drives the inversion of
Section~\ref{sec:inequality:inversion} depends on the local-pass cost $c_L$ being
genuinely small on owned, distributed hardware. A capable forward tier is typically
sparse or partitioned, and if its experts or layer blocks are placed without regard
to the physical tier topology, an ordinary forward pass becomes cross-tier-chatty
and $c_L$ inflates with communication, the very regime R2 tries to avoid, now
imposed on every query rather than only escalations. R3 requires placement that
keeps a forward pass local to its tier, ideally as a trained-in objective so the
model's sparsity structure matches the topology it will be split across.

\noindent
R1 follows from the inequality; R2 and R3 follow from requiring its cost terms be
physically achievable in a tiered deployment rather than nominal. Together they
describe a model whose parameters are aware of the seam, which is precisely what
frozen-checkpoint orchestration, by construction, is not.

\subsection{The primitive gap}
\label{sec:requirements:gap}

None of the three requirements is speculative; each is met, in isolation, by an
existing trained-in primitive (Table~\ref{tab:primitives}). Calibrated early exit
attaches confidence-bearing exit heads inside the network, yielding a per-layer
signal at the cost of the forward pass already in flight \citep{Schuster2022CALM,
CECoLLM2024, Elhoushi2024LayerSkip}; this addresses R1, with the caveat that the
exploitable headroom appears to shrink on the deepest modern models
\citep{DiminishingEarlyExit2026}. Learned activation bottlenecks, developed for
split computing, compress the representation that crosses a partition boundary to a
fraction of its size \citep{Matsubara2022BottleFit, FourierCompress2025}; this
addresses R2. Locality-aware routing objectives train a mixture-of-experts model so
that token-to-expert assignment clusters by locality rather than scattering across
devices \citep{Li2024LocMoEPlus, LSHMoE2024}; this addresses R3, where the
prevailing alternative optimizes placement after training rather than as an
objective \citep{DanceMoE2025}.

\begin{table}[t]
  \centering
  \small
  \begin{tabular}{@{}lccc@{}}
    \toprule
    Trained-in primitive & R1 signal & R2 payload & R3 locality \\
    \midrule
    Calibrated early exit \citep{Schuster2022CALM, CECoLLM2024}      & \checkmark & & \\
    Learned activation bottleneck \citep{Matsubara2022BottleFit}     & & \checkmark & \\
    Locality-aware MoE \citep{Li2024LocMoEPlus}                      & & & \checkmark \\
    \bottomrule
  \end{tabular}
  \caption{Each requirement of Section~\ref{sec:requirements:three} is satisfied in
  isolation by a trained-in primitive; no single trained model satisfies all three.
  The primitives originate in three separate literatures, adaptive-depth
  inference, split computing, and mixture-of-experts systems, developed for
  unrelated ends.}
  \label{tab:primitives}
\end{table}

The pattern in Table~\ref{tab:primitives} is the gap. The diagonal is filled and the
rest is empty: the early-exit models are not built with compressible cross-tier
handoffs or locality-aware placement; the split-computing bottlenecks carry no
calibrated exit signal and no expert-placement objective; the locality-aware
mixtures expose no calibrated escalation signal. The primitives have not been
co-developed because they answer to different communities and different benchmarks.
A model that satisfied all three jointly, a calibrated signal at the seam, a
compressible handoff across it, and placement that keeps the forward pass local, is not a combination any current checkpoint provides. Whether such a model can be
trained, and whether it clears the operating points a tuned frozen cascade cannot,
is the empirical question Section~\ref{sec:architecture} frames as future work.


\section{A candidate architecture and how to test it}
\label{sec:architecture}

What follows is a proposal, not an evaluation. We describe an architecture that
would satisfy R1--R3 jointly, state the hypothesis it embodies in the terms of
Section~\ref{sec:inequality}, and specify a protocol that would test it. We do not
build or evaluate this architecture in this paper;\ifanon\else\ it is the
position's forward-looking conclusion, and\fi\ its validation is left as the
program below.

\subsection{The architecture}
\label{sec:architecture:design}

The three requirements describe one model, not three. A single network is trained
so that a prefix of its layers constitutes the forward-tier model and the whole
constitutes the backstop, with three trained-in features at the layer we have been
calling the seam.

First, a calibrated exit head at the seam (R1) produces both an answer and a
confidence, trained with a calibration-aware objective so that the confidence is
usable as an escalation signal at the cost of the forward pass already computed,
rather than a multiple of it \citep{Schuster2022CALM, CECoLLM2024,
Elhoushi2024LayerSkip}. Second, a learned bottleneck at the seam (R2) compresses
the hidden state into a compact code, trained end-to-end so that the upstream layers
can resume computation from the decompressed code; escalation then transmits the
code and \emph{continues} the forward pass on the backstop rather than re-running it
from the prompt \citep{Matsubara2022BottleFit, FourierCompress2025}. Third, a
locality objective on the model's sparsity (R3) places the prefix experts on the
forward tier and the remainder on the backstop, so that a non-escalated forward pass
stays local and $c_L$ does not inflate with cross-tier traffic
\citep{Li2024LocMoEPlus}. The forward tier runs its prefix locally, emits an answer
and a calibrated confidence at no added cost, and on low confidence escalates a
compressed code from which the backstop resumes. The synthesis is the contribution:
each feature exists, but no current checkpoint carries all three, and the point of
training them together is that they interact at the same seam.

\subsection{What it would change}
\label{sec:architecture:hypothesis}

In the terms of Section~\ref{sec:inequality}, the architecture targets the
$\kappa \to 0$, low-$e_q$ corner of Figure~\ref{fig:feasibility} that a frozen
cascade cannot reach. A frozen cascade reaches that corner only by accident: its
cheap signal has high $e_q$, its calibrated signal pays the $N^\star$ penalty of
Eq.~\eqref{eq:nstar}, and its escalation re-runs from scratch. The co-designed model
is built to occupy the corner, a calibrated seam signal lowers $e_q$ without
multiplying $\kappa$ (R1), a compressed resume-not-rerun handoff keeps the escalation
cheap (R2), and locality keeps $c_L$ small on owned hardware so the deployment sits
at small $\kappa$ (R3). The hypothesis (P4) is that this raises the feasibility
threshold $\kappa^\star = 1 - e_q$ enough to clear operating points a tuned frozen
cascade cannot. Whether the interaction of the three features at one seam is
net-positive, or whether they trade off against one another, is exactly what an
experiment must decide; the inequality predicts the direction of the effect, not its
magnitude.

\subsection{A validation protocol}
\label{sec:architecture:protocol}

The protocol tests P4 against its own falsifier from Section~\ref{sec:propositions}:
that a model meeting R1--R3 fails to improve the quality--cost frontier over a
\emph{tuned} frozen-checkpoint cascade.

\paragraph{Go/no-go gate.}
The first experiment measures the one risk that can end the program early. R1 assumes
a calibrated, accurate exit head is attainable at the seam, and recent evidence
suggests the exploitable early-exit headroom narrows on the deepest models
\citep{DiminishingEarlyExit2026}. Before training the full architecture, we would
measure, at the target scale, whether an exit head at the chosen prefix depth yields
a risk--coverage curve usefully better than the cheap single-pass signals it must
replace. If it does not, R1's primitive does not hold at that scale, and the
architecture's premise fails; the honest outcome is to report that and stop.

\paragraph{Comparison.}
Conditional on the gate, the treatment is the co-designed model and the baseline is
the strongest tuned frozen cascade on a matched parameter budget: the same prefix
model as the small tier and the same full model as the backstop, with a calibrated
router and its threshold swept to trace the baseline frontier \citep{Chen2023FrugalGPT}.
Sweeping the co-designed model's exit threshold traces its frontier. The primary
metric is the empirical analogue of Figure~\ref{fig:feasibility}: whether the
co-designed frontier dominates, lower cost at equal quality, or higher quality at
equal cost, and whether it remains feasible at values of $\kappa$ where the baseline
frontier is empty.

\paragraph{Mechanism measurements.}
To attribute any difference to the three requirements rather than to confounds, we
would separately measure each leg: the calibration of the seam signal as a
risk--coverage curve (R1), the seam payload size and its contribution to escalation
latency (R2), and the cross-tier traffic of a non-escalated forward pass (R3).
Evaluation would span an in-distribution workload \citep{Cobbe2021GSM8K} and a
distribution-shifted one, since the premise under test is signal reliability under
field conditions, where the cheap signals are known to degrade.

\paragraph{Scope.}
This is a focused frontier comparison on a single small/large pair and a small number
of workloads, not a production system and not a frontier-scale training effort. It is
a multi-week study on modest accelerator capacity, and its deliverable is the
frontier comparison above, evidence for or against P4, rather than a deployable
artifact. We flag the cost here so the program is read as falsifiable and bounded,
not open-ended.


\section{Discussion}
\label{sec:discussion}

The feasibility condition changes how the activity around tiered inference reads.
We draw out three consequences and one caution.

\paragraph{Fragmentation as a symptom.}
The routing and cascading literature is large and still growing, and it is usually
read as a healthy diversity of methods, a catalogue of routers, calibrators, and
thresholds, each improving on the last \citep{MoslemKelleher2026RoutingSurvey}.
Proposition~\ref{prop:feasibility} suggests a different reading. Each method is a
way of reaching a lower $e_q$ at a given operating point: a better point inside the
feasible region. None of them moves the region's boundary, because the boundary is
set by $1-\kappa$ and by the frozen-checkpoint structure that fixes how the signal
is obtained, neither of which a new router touches. A field optimizing inside a
boundary it has not characterized will produce exactly this signature, a steady
accumulation of marginal improvements with no theory of when they run out. The
inequality supplies the missing boundary, and with it a way to tell progress toward
the frontier from motion along it.

\paragraph{The inversion redirects the question.}
Decentralized inference is normally argued as a cost, latency, or data-residency
play. The $\kappa$ analysis recasts it as a change in which lever moves the outcome. On the
sunk-cost edge, where $\kappa \to 0$, the cost ceiling that
\citet{Bouchard2026Escalation} finds dominant in the paid regime stops excluding
operating points, and signal quality instead sets the fraction of the workload the
deployment resolves locally (Section~\ref{sec:inequality:inversion}). The operative
research question for decentralized inference is therefore not how to route cheaply (cost is already slack) but how to obtain a calibrated escalation signal at the
seam, which is R1. The same move that makes decentralization attractive on cost
makes signal quality the thing that decides whether it works. Effort spent driving
down routing cost in this regime is spent against a constraint that no longer binds.

\paragraph{Structural change, not signal engineering.}
The one form of tiering that consolidated into a production standard did so by
sidestepping the confidence signal entirely: phase disaggregation splits by
structure, not by difficulty, and is limited by a quantity, interconnect bandwidth, that does not depend on a model's self-knowledge \citep{Zhong2024DistServe}. The
lesson generalizes. Durable gains in tiered inference have come from structural
changes to how computation is divided, not from better estimates of a frozen model's
reliability. The co-designed architecture of Section~\ref{sec:architecture} is that
lesson applied to the difficulty-routing problem: rather than bolt a smarter router
onto a frozen checkpoint, build the tier boundary into the model. Whether that
succeeds is open, but the direction follows from where the boundary actually sits.

\paragraph{A caution on where the margin is measured.}
The feasibility margin $1-\kappa-e_q$ is most favorable where the signal is best,
and the signal is best in distribution. Reported cascade results are typically
measured on benchmarks drawn from the same distribution as the router's tuning data;
field deployment is not, and the cheap signals degrade under shift
\citep{Wen2025KnowYourLimits, Kotte2026UCCI}. A deployment that appears comfortably
inside the feasible region on a benchmark can sit on its boundary in the field, as
$e_q$ rises with the signal's degradation. This is not a flaw in the condition but a
warning about the inputs it is evaluated on, and it is the reason the protocol of
Section~\ref{sec:architecture:protocol} tests under distribution shift rather than
in distribution alone. Section~\ref{sec:limitations} turns to the assumptions on
which the condition itself rests.


\section{Limitations}
\label{sec:limitations}

We separate limitations of three kinds: assumptions inside the formal model,
exposure carried by the architectural proposal, and the evidential base the
argument rests on. For each assumption in the model we state the direction of the
effect when it is relaxed, because several of them are conservative (the realistic case tightens the ceiling rather than loosening it), and a limitations section that
did not say so would understate the result as readily as overstate it.

\subsection{Assumptions in the formal model}
\label{sec:limitations:model}

\paragraph{The cost model is generate-then-judge.}
Proposition~\ref{prop:feasibility} charges every query the forward-pass cost $c_L$
because the small model must run to produce the confidence signal
(Section~\ref{sec:inequality:constraints}). This is the cascade case, and it is the
case the paper is about: a signal read from the forward model's own output. A
\emph{predictive} router that decides before running the small model, from features
of the query alone, has a different cost structure and is not covered by this form
of the condition. We regard this as a scope boundary rather than a threat, with one
regime-dependent observation: a pre-generation router forgoes the forward model's
output as a signal source and must route on weaker query-level features, but it also
avoids paying $c_L$ on the queries it sends upstream. At small $\kappa$, where that
saving is negligible, the cascade case we analyze is the one more favorable to
tiering; at appreciable $\kappa$ the router's cost advantage can dominate its weaker
signal, and \citet{Bouchard2026Escalation} reports exactly this inversion
empirically.

\paragraph{The backstop's error is taken independent of the escalated set.}
We set the backstop error to a constant $u$ regardless of which queries are
escalated. In deployment the escalated queries are the ones the forward model found
hard, and a capable backstop is often \emph{also} weaker on them, so the realistic
error on the escalated set is some $u' \ge u$. Substituting $u'$ for $u$ in the
escalated-error term of Eq.~\eqref{eq:cascade-error} raises $E_{\text{cascade}}$,
which raises $e_q$ and shrinks the feasible region: the constant-$u$ assumption is
optimistic, and relaxing it \emph{strengthens} the impossibility. We sketch the
monotonicity here and give the formal statement in
Appendix~\ref{sec:appendix-derivation}. \emph{Confidence: high} that the direction is
as stated; the magnitude depends on the correlation between the two models'
difficulty, which is workload-specific.

\paragraph{The analysis is two-tier.}
The condition concerns a single seam between a forward tier and a backstop, while
deployments chain several tiers (local, edge, regional, datacenter). We do not prove
the multi-tier case. We expect it does not loosen the two-tier result: each seam in a
chain is a two-tier decision subject to its own instance of
Proposition~\ref{prop:feasibility}, and a chain is feasible only where every seam is,
so the multi-tier feasible region is an intersection of two-tier regions and no
larger than the tightest. \emph{Confidence: moderate}; this is a composition
argument, not a derivation, and cross-tier interactions we have not modeled could
complicate it.

\paragraph{Only the extremes of router quality are solved in closed form.}
We evaluate $e_q$ exactly at the chance and perfect operating points
(Corollaries~\ref{cor:floor} and~\ref{cor:ceiling}), which bracket the feasible
range of $\kappa$. The contingent band between them
(Section~\ref{sec:inequality:ceiling}) is defined operationally; its exact shape
depends on the router's ROC parameterization, which we do not commit to. The bracket
is what the argument needs (P2 and P3 follow from the two thresholds), but a reader wanting the middle curve for a specific signal would supply an ROC model (for
example binormal) that we leave open.

\paragraph{Cost is a single scalar, and latency is folded into it.}
The ratio $\kappa$ collapses money, energy, throughput, and latency into one number.
Where one of these dominates, the scalar is adequate; where two trade off, as when wide-area round-trip latency matters independently of compute cost (the stop-and-wait coordination of selective cross-tier speculative decoding \citep{EdgeCloudSpecDecode2026}), a single $\kappa$ does not capture the deployment,
and a multi-objective version of the condition would be needed. This is a genuine
restriction of scope, not a conservative assumption, and we flag it as such.

\paragraph{Sunk cost is an idealization, not literally zero.}
The inversion is stated at $\kappa \to 0$, but owned hardware still carries amortized
capital, energy, and opportunity cost, so a real edge deployment sits at a small but
positive $\kappa$. The argument does not need the limit: the governing quantity
shifts toward signal quality across a neighborhood of small $\kappa$, not only at the
point $\kappa = 0$. We use the limit for clarity, not as a claimed operating point.

\subsection{The architectural proposal}
\label{sec:limitations:proposal}

The candidate architecture of Section~\ref{sec:architecture} is a hypothesis, and its
three legs are not equally grounded. R1 (calibrated early exit) and R3
(locality-aware placement) map onto established primitives; R2 as we use it, training a bottleneck so the backstop \emph{resumes} a partially computed forward pass
from a compressed seam code, rather than merely compressing a transmitted activation, goes beyond what the split-computing literature establishes
\citep{Matsubara2022BottleFit}. Whether a backstop can resume from such a code across
the seam, with a different upper stack, is part of what the protocol of
Section~\ref{sec:architecture:protocol} would test, not something we assume. The R2
leg therefore carries more unvalidated weight than the other two, and we mark it as
the architectural bet most likely to fail.

\subsection{The evidential base}
\label{sec:limitations:evidence}

The argument is motivated by a finding we cite rather than establish: that cheap,
single-pass confidence signals are miscalibrated and brittle, and that the reliable
alternative is expensive \citep{Wen2025KnowYourLimits, Kotte2026UCCI}. If that
finding were wrong, or held only on narrow workloads, the motivation for the whole
analysis would weaken: a well-calibrated cheap signal would keep $e_q$ low without
self-consistency, and the squeeze of Eq.~\eqref{eq:nstar} would not bite. We rely on
it because it is replicated across the abstention and calibration literature, but it
is an input to our argument, not an output of it, and the protocol's
distribution-shift arm (Section~\ref{sec:architecture:protocol}) is in part a test of
whether it holds where deployments actually run.

\medskip
\noindent
The net picture: the cost-model scope, the two-tier restriction, and the scalar
$\kappa$ bound what the condition speaks to; the constant-$u$ assumption is
conservative and its relaxation tightens the result; and the load-bearing exposures
are the R2 architectural bet and the cited signal-brittleness finding, both of which
the proposed protocol is designed to probe rather than presume.


\section{Ethics and broader impact}
\label{sec:ethics}

Tiered inference is argued, in part, on privacy: keep computation on the user's
device and escalate only when necessary, so that most data never leaves the local
tier. The mechanism this paper analyzes has a consequence that cuts against that
framing, and the consequence is structural rather than incidental.

\paragraph{The privacy inversion.}
The escalation criterion is the forward model's low confidence, and a model is least
confident on inputs that are atypical for it: this is the basis of selective
prediction itself \citep{Geifman2017Selective}. The fraction of traffic that crosses
the seam to the datacenter is therefore not a random sample of queries; it is
selected for atypicality. Atypical queries are disproportionately the unusual, rare,
long-tail cases, and sensitivity tends to live in that same tail: the unusual medical
situation, the identifying personal detail, the edge-case circumstance a common query
does not contain. A system that keeps the mundane local and escalates the atypical
thus ships the more sensitive fraction of traffic off-device while retaining the
benign remainder, close to the opposite of the privacy guarantee the local-first
framing suggests. We state the correlation between atypicality and sensitivity as a
reasoned expectation, not a measured fact; it is itself an empirical claim, and one a
deployment should verify rather than assume. \emph{Confidence: moderate.}

\paragraph{Who escalates.}
The forward model's accuracy $q$ is not uniform across users. A small model is
typically weaker on dialects, minority languages, and underrepresented domains, so
queries from those users fall below the confidence threshold more often and escalate
more often. The privacy cost of data leaving the device, and the latency cost of the
round trip, therefore fall most heavily on the users the small model already serves
worst on accuracy. The same non-uniformity in $q$ that lowers their answer quality
raises their escalation rate: worse accuracy, weaker privacy, and higher latency
concentrate on the same population. This is a distributional consequence of routing
on a signal whose quality varies across groups, and it is invisible to an aggregate
metric that reports only the mean escalation rate.

\paragraph{Mitigations, and their tensions.}
The architecture of Section~\ref{sec:architecture} inherits these consequences and,
because it is co-designed, is also where they could be addressed, though not without
tension. The learned bottleneck of R2 could be trained to suppress identifying detail
in the escalated code, but the backstop needs enough of the representation to answer,
so a privacy-preserving bottleneck trades against the quality the escalation exists to
provide; this is an open problem, not a solved one. Per-group escalation rates can be
made a first-class deployment metric, surfacing the disparity that the mean hides, but
measuring it requires the group labels that privacy-conscious deployments avoid
collecting. A user can be allowed to refuse escalation for a sensitive query and accept
the local model's answer, but that returns the user to the lower-accuracy tier exactly
when the query is hardest. We raise these not as fixes but as the design space the
inversion opens, and as reasons the privacy case for tiered inference should be made on
the escalated fraction, not the local one.

\paragraph{Scope.}
None of this is an argument against tiered or decentralized inference, which has real
privacy and efficiency benefits. It is an argument that those benefits are not
automatic, that they accrue to the queries that stay local and not to the ones that
leave, and that the selection of which queries leave is correlated with both
sensitivity and group membership in ways a deployment should measure. The point is to
design and report against the consequence, not to abandon the architecture that
produces it.


\section{Future work}
\label{sec:future}

The central direction is the validation program of
Section~\ref{sec:architecture:protocol}: train a model that meets R1--R3 jointly and
test, against a tuned frozen-checkpoint cascade, whether it clears operating points
the cascade cannot (P4). That experiment is the paper's natural sequel, and its
go/no-go gate, whether usable early-exit headroom survives at the target scale, is the first thing it would resolve.

Three extensions would strengthen the formal result. The condition is two-tier;
generalizing it to a chain of tiers requires either proving the composition argument
of Section~\ref{sec:limitations:model}, that a multi-tier feasible region is the
intersection of its per-seam two-tier regions, or finding the cross-tier interaction
that breaks it. The cost ratio is a scalar; a multi-objective form that separates
compute cost from wide-area latency would cover the deployments, such as selective
cross-tier speculative decoding, where round-trip time binds independently of compute
\citep{EdgeCloudSpecDecode2026}. And the contingent band between the chance and
perfect thresholds is defined operationally; instantiating $e_q$ with an explicit ROC
model would turn the bracket of Section~\ref{sec:inequality:ceiling} into a full
feasibility surface, at the cost of committing to a parameterization the argument does
not currently need.

Two of the paper's claims are empirical and unsettled, and testing them matters more
than extending the algebra. The feasibility margin is evaluated where the signal is
best, in distribution, and how far it shrinks in the field, as $e_q$ rises under
distribution shift, is measurable and, to our knowledge, unmeasured for tiered
deployments. The distributional consequences of
Section~\ref{sec:ethics} are likewise testable: whether escalation rates are in fact
higher for the users a small model serves worst, and whether the atypicality that
triggers escalation correlates with sensitivity as we expect, are questions a
deployment can answer with per-group instrumentation. The argument depends on the
first of these and implies the second; both deserve measurement rather than the
reasoned expectation we have offered.


\section{Conclusion}
\label{sec:conclusion}

We began with a question about tiered inference: whether pushing computation outward
relocates the constraint that binds, or only relocates the compute. The answer is
that it depends on a single quantity the literature had left implicit. Modeling the
routing decision as a binary test and denominating local and escalation cost through
the architecture cost ratio $\kappa$, we derived a closed-form condition for when no
threshold on a confidence signal makes a tiered cascade worthwhile
(Proposition~\ref{prop:feasibility}), bracketed by a signal-free floor below which a
cascade works at any signal quality and a perfect-signal ceiling above which none can
(Corollaries~\ref{cor:floor} and~\ref{cor:ceiling}). The four mechanisms that make up
the current toolkit share the dependency the condition is about, and the one that
escaped it, phase disaggregation, is the one that reached production. The
condition turns into design requirements that existing trained-in primitives meet
only in isolation, never jointly.

The condition's most consequential reading is along the $\kappa$ axis. Where local
computation is billed, cost binds, as recent work finds; where it is sunk, on the
owned hardware of a decentralized deployment, feasibility becomes unconditional and
signal quality instead sets what the cascade achieves. Decentralization is therefore
not a cost optimization but a change in which lever moves the outcome, and the
research question it raises is not how to
route cheaply but how to obtain a calibrated signal at the seam. Because the boundary
is set by frozen-checkpoint orchestration, moving it calls for a model co-designed
around the tier it will be split across rather than a better router bolted onto a
frozen one. That architecture, and the program to test it, are the paper's
forward-looking conclusion and remain a hypothesis. The condition and the inversion do
not: they hold whether or not the architecture succeeds, and they are what we ask the
reader to take from this paper.

\appendix

\section{Full derivation of the feasibility condition}
\label{sec:appendix-derivation}

This appendix reproduces in full the steps shown in outline in
Section~\ref{sec:inequality}, and proves the two claims deferred to it: that the
signal-free floor never exceeds the perfect-signal ceiling
(Lemma~\ref{lem:ordering}), and that relaxing the constant-backstop-error assumption
tightens the ceiling (Lemma~\ref{lem:monotone}). Notation is that of
Table~\ref{tab:notation}; we work in the non-degenerate regime
\begin{equation}
  (1-q) > u, \qquad 0 < \delta < (1-q)-u,
  \label{eq:appendix-regime}
\end{equation}
so that the forward model errs more often than the backstop (escalation can reduce
error) and the tolerance is smaller than the gap to be closed (the cascade is not
satisfied by the forward model alone).

\subsection{Cost constraint}
Every query is run on the forward model at cost $c_L$; the escalated fraction $e$ is
additionally run on the backstop at cost $c_D$. The cascade cost per query is
$c_L + e\,c_D$, and the all-to-backstop baseline is $c_D$. The cascade is cheaper iff
\begin{equation}
  c_L + e\,c_D < c_D
  \;\Longleftrightarrow\;
  c_L < (1-e)\,c_D
  \;\Longleftrightarrow\;
  \kappa < 1 - e
  \;\Longleftrightarrow\;
  e < 1 - \kappa,
  \label{eq:appendix-cost}
\end{equation}
dividing by $c_D > 0$ and using $\kappa = c_L/c_D$.

\subsection{Error decomposition and quality constraint}
Partition the workload by whether the forward model $L$ is correct and whether the
router escalates. The four cells, their probabilities, the answer returned, and the
error contributed are:
\begin{center}
\begin{tabular}{@{}llll@{}}
\toprule
$L$ correct? & escalate? & probability & answer / error contribution \\
\midrule
yes & no  & $q(1-f)$     & $L$, correct $\Rightarrow 0$ \\
yes & yes & $qf$         & $D$, error $u \Rightarrow qf\,u$ \\
no  & no  & $(1-q)(1-s)$ & $L$, wrong $\Rightarrow (1-q)(1-s)$ \\
no  & yes & $(1-q)s$     & $D$, error $u \Rightarrow (1-q)s\,u$ \\
\bottomrule
\end{tabular}
\end{center}
Summing the error column and using $e = qf + (1-q)s$,
\begin{equation}
  E_{\text{cascade}}
  = (1-q)(1-s) + u\big[qf + (1-q)s\big]
  = (1-q)(1-s) + u\,e.
  \label{eq:appendix-error}
\end{equation}
Requiring the cascade error to exceed the baseline $u$ by at most $\delta$ gives the
quality constraint
\begin{equation}
  (1-q)(1-s) + u\,e \le u + \delta.
  \label{eq:appendix-quality}
\end{equation}

\subsection{Feasibility}
For a router with a fixed ROC curve, sliding the decision threshold traces operating
points $(f,s)$ and hence escalation rates $e$. Let $e_q$ be the smallest escalation
rate at which the router, at its best operating point for that rate, satisfies
\eqref{eq:appendix-quality}. (The set of quality-satisfying $e$ is an interval; $e_q$
is its left endpoint.)

\begin{proposition}[restatement of Proposition~\ref{prop:feasibility}]
A cascade admits an operating point satisfying both \eqref{eq:appendix-cost} and
\eqref{eq:appendix-quality} if and only if $\kappa < 1 - e_q$.
\end{proposition}

\begin{proof}
Cost requires $e < 1-\kappa$; quality requires $e \ge e_q$. The two are jointly
satisfiable iff the interval $[e_q,\,1-\kappa)$ is nonempty, i.e.\ iff
$e_q < 1-\kappa$, i.e.\ iff $\kappa < 1 - e_q$. The smallest quality-feasible $e$ is
$e_q$, so picking $e$ just above $e_q$ (and below $1-\kappa$) realizes a feasible
operating point whenever the inequality holds.
\end{proof}

\subsection{The two thresholds}
Evaluating $e_q$ at the extremes of router quality gives the bracketing thresholds.

\paragraph{Chance signal (floor).}
An uninformative signal lies on the ROC diagonal $s=f$, whence $e = s(1-q)+sq = s$.
Substituting $s=e$ into \eqref{eq:appendix-quality},
\begin{align}
  (1-q)(1-e) + u\,e &\le u + \delta \notag\\
  (1-q) - u - \delta &\le e\big[(1-q) - u\big] \notag\\
  e \;\ge\; e_q^{\mathrm{ch}} &= \frac{(1-q)-u-\delta}{(1-q)-u} = 1 - \frac{\delta}{(1-q)-u},
  \label{eq:appendix-eq-chance}
\end{align}
using $(1-q)-u>0$. Then $\kappa < 1 - e_q^{\mathrm{ch}} = \delta/[(1-q)-u] =
\kappa^\star_{\mathrm{floor}}$, and the feasible set is empty under an uninformative
signal exactly when $\kappa\,[(1-q)-u] \ge \delta$.

\paragraph{Perfect signal (ceiling).}
A perfectly discriminating signal escalates only the forward model's errors ($f=0$),
so for an escalation budget $e \le 1-q$ it catches a fraction $s = e/(1-q)$ of them
and $(1-q)s = e$. Then $(1-q)(1-s) = (1-q)-e$, and \eqref{eq:appendix-quality} becomes
\begin{align}
  (1-q) - e + u\,e &\le u + \delta \notag\\
  (1-q) - u - \delta &\le e\,(1-u) \notag\\
  e \;\ge\; e_q^{\mathrm{pf}} &= \frac{(1-q)-u-\delta}{1-u}.
  \label{eq:appendix-eq-perfect}
\end{align}
The requirement $e_q^{\mathrm{pf}} \le 1-q$ holds throughout
\eqref{eq:appendix-regime}: it reduces to $-qu \le \delta$, true for $\delta>0$. Then
$\kappa < 1 - e_q^{\mathrm{pf}} = \kappa^\star_{\mathrm{ceil}}$.

\begin{lemma}[Ordering]
\label{lem:ordering}
Under \eqref{eq:appendix-regime}, $\kappa^\star_{\mathrm{floor}} \le
\kappa^\star_{\mathrm{ceil}}$; equivalently $e_q^{\mathrm{pf}} \le e_q^{\mathrm{ch}}$.
\end{lemma}

\begin{proof}
Write $a = (1-q)-u > 0$. From \eqref{eq:appendix-eq-chance} and
\eqref{eq:appendix-eq-perfect}, $e_q^{\mathrm{ch}} = (a-\delta)/a$ and
$e_q^{\mathrm{pf}} = (a-\delta)/(1-u)$, with $a-\delta>0$ by
\eqref{eq:appendix-regime}. Dividing, $e_q^{\mathrm{pf}}/e_q^{\mathrm{ch}} =
a/(1-u)$, and $a = (1-q)-u \le 1-u$ since $q \ge 0$. Hence $e_q^{\mathrm{pf}} \le
e_q^{\mathrm{ch}}$, and $1-e_q$ reverses the inequality.
\end{proof}

For $\kappa^\star_{\mathrm{floor}} \le \kappa < \kappa^\star_{\mathrm{ceil}}$
feasibility is contingent on signal quality; outside this band it is settled in one
direction or the other.

\subsection{Self-consistency}
A self-consistency signal of $N$ samples multiplies the local cost,
$c_L \mapsto N c_L$, hence $\kappa \mapsto N\kappa$. Feasibility
($N\kappa < 1-e_q$) holds iff $N < N^\star = (1-e_q)/\kappa$. Taking the signal-free
floor as reference, $1-e_q^{\mathrm{ch}} = \delta/[(1-q)-u]$, so
\begin{equation}
  N^\star = \frac{\delta}{\kappa\,[(1-q)-u]} = \frac{\delta\,\rho}{(1-q)-u},
  \qquad \rho = 1/\kappa.
  \label{eq:appendix-nstar}
\end{equation}

\subsection{Monotonicity in backstop difficulty}
We now relax the assumption that the backstop error is the constant $u$ regardless of
which queries escalate (Section~\ref{sec:limitations:model}). Let $u_e$ be the
backstop's error on the \emph{escalated} subset; the baseline error over the whole
workload remains $u$. Only escalated queries are answered by the backstop, so $u$ is
replaced by $u_e$ in \eqref{eq:appendix-error}, while the baseline on the right of
\eqref{eq:appendix-quality} stays $u$:
\begin{equation}
  (1-q)(1-s) + u_e\,e \le u + \delta.
  \label{eq:appendix-quality-ue}
\end{equation}

\begin{lemma}[Monotonicity]
\label{lem:monotone}
For a fixed router, the minimum quality-feasible escalation rate $e_q$ is
non-decreasing in $u_e$ on $u_e \in [u,\,1-q)$. Consequently the feasibility
threshold $\kappa^\star = 1 - e_q$ is non-increasing in $u_e$, and the constant-error
case $u_e = u$ is the most permissive: any $u_e > u$ weakly shrinks the feasible
region.
\end{lemma}

\begin{proof}
Take the chance router ($s=e$); the argument for a general fixed ROC is identical
term by term. From \eqref{eq:appendix-quality-ue} with $s=e$,
\[
  (1-q) - (1-q)e + u_e e \le u + \delta
  \;\Longleftrightarrow\;
  e \ge \frac{(1-q)-u-\delta}{(1-q)-u_e} \equiv e_q(u_e),
\]
valid while $(1-q)-u_e > 0$. The numerator $(1-q)-u-\delta > 0$ is fixed by
\eqref{eq:appendix-regime} and independent of $u_e$; the denominator
$(1-q)-u_e$ is positive and strictly decreasing in $u_e$. Hence $e_q(u_e)$ is
non-decreasing in $u_e$, with $e_q(u) = e_q^{\mathrm{ch}}$ recovering
\eqref{eq:appendix-eq-chance} and $e_q(u_e) \to \infty$ as $u_e \to (1-q)^-$ (the
backstop ceases to help, and no escalation budget suffices). Since
$\kappa^\star = 1 - e_q$, it is non-increasing in $u_e$.
\end{proof}

Because escalated queries are those the forward model found hard, and a capable
backstop is typically also weaker on them, the realistic case is $u_e > u$. By
Lemma~\ref{lem:monotone} the constant-$u$ analysis of the body is therefore
optimistic, and the impossibility results of Section~\ref{sec:inequality} hold a
fortiori under the realistic relaxation.

\subsection{Worked values}
For $q = 0.6$, $u = 0.05$, $\delta = 0.02$, so $(1-q)-u = 0.35$:
\eqref{eq:appendix-eq-chance} gives $e_q^{\mathrm{ch}} = 1 - 0.02/0.35 \approx 0.943$
and $\kappa^\star_{\mathrm{floor}} \approx 0.057$;
\eqref{eq:appendix-eq-perfect} gives $e_q^{\mathrm{pf}} = 0.33/0.95 \approx 0.347$ and
$\kappa^\star_{\mathrm{ceil}} \approx 0.653$, consistent with
Lemma~\ref{lem:ordering}; and \eqref{eq:appendix-nstar} at $\rho = 30$ gives
$N^\star = 0.6/0.35 \approx 1.7$.

\bibliographystyle{tmlr}
\bibliography{references}

\end{document}